\documentclass[runningheads]{llncs}

\usepackage{eccv}

\usepackage{eccvabbrv}

\usepackage{graphicx}
\usepackage{booktabs}

\usepackage{xcolor,pifont}
\newcommand{\cmark}{\textcolor{green!60!black}{\ding{51}}}
\newcommand{\xmark}{\textcolor{red!70!black}{\ding{55}}}
\usepackage{multirow}

\usepackage{tabularx}
\usepackage{makecell}
\usepackage{array}

\usepackage[accsupp]{axessibility}  

\usepackage{hyperref}

\usepackage{orcidlink}

\begin{document}

\title{Low-Rank Ternary Adaptation for Fine-Tuning Transformers}


\author{Alexandru-Dragos Manolache\inst{1}\textsuperscript{*}\orcidlink{0009-0005-3456-0334}
\and
Yunqiang Li\inst{2}\textsuperscript{*}
\and
Jan van Gemert\inst{1}\orcidlink{0000-0002-3913-2786}}

\authorrunning{A.~Manolache et al.}

\institute{Delft University of Technology, The Netherlands \and
Amazon Development Center
}

\maketitle

\begingroup
\renewcommand{\thefootnote}{*}
\footnotetext{Equal contribution.}
\endgroup

\begin{abstract} 
Ternary transformers offer extreme memory and compute efficiency,
but existing low-bit LoRA-based methods cannot directly fine-tune ternary weights. Current approaches
either require dequantization, restoring low-bit base weights to higher precision to merge with adaptation weight, or update only quantization parameters, preventing a merged model that remains ternary. We propose ternary multiplicative adaptation, which represents discrete updates of ternary weights such as sign flips or zeroing through a low-rank Kronecker factorization into two small ternary matrices
applied element-wise to ternary weights. This design is parameter-efficient and expressive, preserves the ternary domain, and supports direct merging without dequantization. Experiments on six models across language and vision, including ternarized LLaMA-3 1B and 3B and a ternary ViT-B/16, demonstrate that our method recovers much of the performance lost to quantization and outperforms strong low-bit and ternary baselines.
Code is available at \url{https://github.com/alexmanoo/ternary_adaptation}.
\end{abstract}
\section{Introduction}
\label{sec:intro}

\begin{table}[ht]
\centering
\small
\setlength{\tabcolsep}{0.9pt}
\renewcommand{\arraystretch}{1.1}
\captionof{table}{ Comparison of QLoRA, QA-LoRA, and our ternary multiplicative adapter.
Each column indicates whether a method satisfies key properties needed for adapting ternary models.
\emph{Base weights ternary} asks if the backbone can be stored in a ternary format during fine-tuning.
\emph{Dequant-free merge} asks whether the adaptation can be merged directly into the base-weights while keeping the low-bit quantization.
\emph{Ternary merged model} indicates whether the final weights remain ternary after merging.
The bottom rows describe how the adaptation is applied: QLoRA adds a low-rank update to dequantized floating-point weights, QA-LoRA adds a low-rank update to quantization parameters, and our method applies a multiplicative ternary adaptation directly to ternary weights. Only our method both operates natively on ternary backbones and produces a merged model that stays ternary while avoiding any dequantization step.}
\begin{tabular}{lccc}
\toprule
\textbf{Property}     & \textbf{QLoRA}  & \textbf{QA-LoRA} & \textbf{Ours} \\
\midrule
Base weights ternary  & \cmark          & \xmark           & \cmark \\
Dequant-free merge    & \xmark          & \cmark           & \cmark \\
Ternary merged model  & \xmark          & \xmark           & \cmark \\
\midrule
Update type           & Additive    & Additive        & Multiplicative \\
Applied on            & FP weights  & Quantizer       & Ternary weights \\
\bottomrule
\end{tabular}
\label{tab:methods-comparison}
\end{table}

\begin{figure*}[t]
  \centering

  \begin{minipage}{0.95\linewidth}
    \centering
    \textbf{Ternary Multiplicative Update}\par\vspace{0.8em}
    \includegraphics[width=\linewidth]{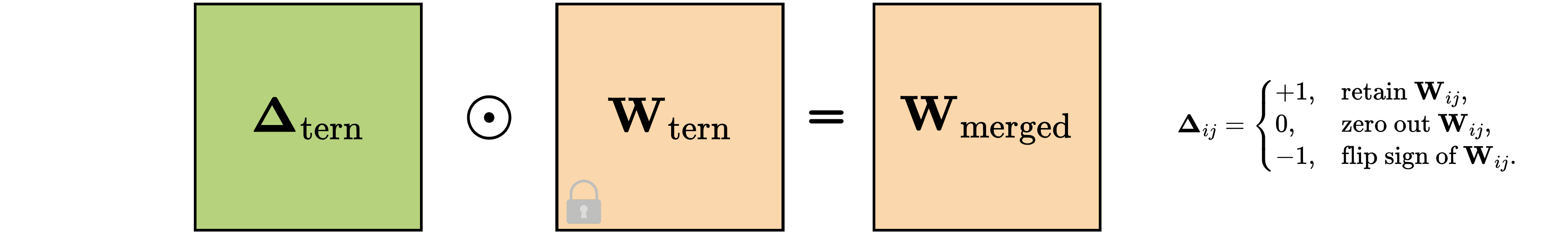}
  \end{minipage}

  \vspace{2.0em}

  \begin{minipage}{0.95\linewidth}
    \centering
    \textbf{Kronecker-structured Ternary Adaptation}\par\vspace{0.8em}
    \includegraphics[width=\linewidth]{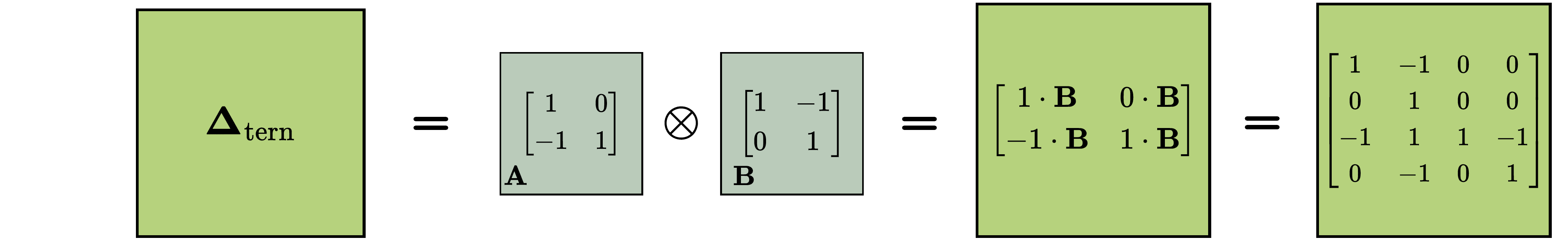}
  \end{minipage}

  \caption{
  Visualization of our ternary multiplicative adapter.
  \emph{Ternary Multiplicative Update} (top): a ternary mask
  $\mathbf{\Delta}_{\text{tern}}\in\{-1,0,1\}^{d_{\text{out}}\times d_{\text{in}}}$
  is applied element-wise (Hadamard product) to the pretrained ternary
  weights $\mathbf{W}_{\text{tern}}$ to obtain merged ternary weights
  $\mathbf{W}_{\text{merged}}$, where each entry
  $\mathbf{\Delta}_{ij}\in\{-1,0,1\}$ respectively keeps, zeros, or flips the sign
  of $\mathbf{W}_\text{ij}$. 
  \emph{Kronecker-structured Ternary Adaptation} (bottom): the adaptation
  $\mathbf{\Delta}_{\text{tern}}$ is constructed as a Kronecker product
  $\mathbf{A}\otimes\mathbf{B}$ of two small ternary factors; the toy
  example shows how each entry of $\mathbf{A}$ scales a copy of
  $\mathbf{B}$ to form a larger ternary matrix. Our adapter uses a compact
  Kronecker-structured mask to implement discrete keep/zero/flip
  operations directly on ternary weights, enabling a mergeable
  fine-tuned model that remains fully ternary.
  }
  \label{fig:ternary_adapter}
\end{figure*}

Transformer models deliver strong performance across many tasks, including language and vision, but their memory and compute requirements make deployment expensive. Quantization offers a solution to reduce these costs by compressing model weights from full-precision to a few bits while preserving most of the accuracy. In current LLMs, 2-bit weight quantization is a viable setting for aggressive compression~\cite{chee2023quip,frantar2022gptq}. In this work, we investigate the next step in compression: ternary quantization which restricts weights to $\{-1, 0, 1\}$ with the advantage of memory savings and throughput efficiency suitable for lower end hardware~\cite{li2016ternary,wang2023bitnetscaling1bittransformers,ma2024era}. The memory budget of a ternary quantized model corresponds to $\log_2 3 \approx 1.58$ bits per weight, reducing the theoretical memory footprint of a 16-bit precision Llama 8B\cite{dubey2024llama} model from $16\text{GB}$ to only $1.6\text{GB}$.

Fine-tuning all weights of large pretrained models is often prohibitively expensive, motivating parameter efficient fine-tuning (PEFT) that adjusts a large network by updating a smaller set of parameters.
The leading PEFT method is Low-Rank Adaptation (LoRA)~\cite{hu2022lora}; it fine-tunes a low-dimensional weight decomposition 
which during inference can be added to the original base weight to obtain an adapted, merged model.
Building on LoRA, recent work has explored PEFT for quantized base models.
QLoRA~\cite{dettmers2023qlora} combines post-training quantization (PTQ) of pre-trained weights with low-rank adaptation. QA-LoRA~\cite{xu2023qa} improves further by integrating the adaptation directly into the quantization parameters. These two methods make LoRA-style adaptation a compelling strategy for low-bit transformer backbones.

However, adapting low-bit quantized models introduces new challenges. QLoRA \cite{dettmers2023qlora} requires dequantizing base weights back to full precision to add with adaptations, after which the merged model must be re-quantized for low-bit inference. QA-LoRA shows that such re-quantization leads to noticeable accuracy degradation at 2-bit widths and therefore proposes a direct merging strategy that integrates the adaptation into per-group quantization parameters, avoiding de-quantization. While effective at 2 bits, QA-LoRA is incompatible with ternary quantization, where both the base weights and the merged weights are constrained to $\{-1, 0, 1\}$. This gap motivates an adaptation design that supports direct merging for ternary models.

In the ternary quantization setting, the update space is discrete. Weights cannot be adjusted by small continuous steps, because the main degrees of freedom are flipping signs or zeroing weights. Conventional LoRA-style approaches~\cite{hu2022lora,dettmers2023qlora,zhang2023adalora,lialin2023relora}, which use continuous additive updates in floating-point space, are therefore not directly compatible with ternary weights for efficient inference~\cite{frantar2022gptq, lin2024awq, shao2023omniquant}. Moreover, representing such discrete ternary updates via trainable low-rank decomposition matrices is non-trivial, as the update space is both discrete and non-linear. A possibility is to temporarily dequantize ternary weights to higher precision, apply a continuous adaptation, and then requantize. However, as observed in prior work at 2 bits~\cite{xu2023qa}, this dequantize-requantize process introduces additional quantization error, which becomes even more harmful for sub-2-bit quantization. We therefore seek a PEFT solution that (i) operates natively in the ternary domain by expressing flips and zeroing of weights via a low-rank representation, and (ii) can be merged into the base ternary weights without any dequantization or requantization.

We tackle two challenges: representing discrete ternary updates, 
and decomposing them with low-rank matrices in a compact yet expressive form.
To this end, we introduce a ternary multiplicative adaptation  
that directly models ternary updates, as illustrated in Figure~\ref{fig:ternary_adapter} and Table~\ref{tab:methods-comparison}, where ternary weights are flipped or zeroed instead of updated continuously. The adaptation is constructed as the Kronecker product of two smaller
ternary matrices and applied via element-wise Hadamard multiplication. 
This design achieves high rank using two trainable low-rank matrices, keeps the number of adaptation parameters small, 
preserves the ternary domain, and implements three operations per weight: keep, zeroing, or flip sign, without leaving the domain. As in prior LoRA-style methods, the adaptation 
can be fully merged into the base model post-tuning. In our case, the merged weights remain ternary, so no dequantization or re-quantization is required. 

We make the following contributions:
1) A ternary multiplicative adaptation that directly models discrete ternary updates for ternary LLMs and vision transformers.
2) A low-rank Kronecker decomposition of the ternary updates, achieving high expressivity while remaining parameter-efficient.
3) The adapter can be fully merged into the base ternary model, enabling zero-overhead deployment. 
4) We evaluate our method on six models across language and vision. On ternarized Llama-3.2 1B and 3B models, our method recovers much of the performance lost to quantization, outperforms stronger 2-bit baselines on most tasks, and outperforms requantized QLoRA baseline under the same 1.58-bit quantization level. We also validate on a ternary ViT-B/16 fine-tuned on ImageNet-100~\cite{ambityga_imagenet100_kaggle}, where our merged 1.58-bit model outperforms requantized QLoRA and narrows the gap to full ternary fine-tuning.

\section{Related Work}
\label{sec:relatedwork}

\paragraph{Quantization in Transformers.}
Quantization compresses model weights and activations into low-bit formats, reducing memory and computation for efficient model deployment~\cite{wei2024advances,du2024model}.  
Post-training quantization (PTQ) methods such as LLM.int8()~\cite{dettmers2022gpt3}, GPTQ~\cite{frantar2022gptq}, AWQ~\cite{lin2024awq}, and OmniQuant~\cite{shao2023omniquant} minimize quantization error through calibration or activation-aware scaling, while SmoothQuant~\cite{xiao_smoothquant_2024} and SpinQuant~\cite{liu2024spinquant} rebalance weight–activation distributions.  
For vision transformers (ViTs)~\cite{dosovitskiy2020image}, PTQ4ViT~\cite{yuan2022ptq4vit} uses activation quantization with Hessian-guided scale calibration, and I-ViT~\cite{li2023vit} targets integer-only ViT inference via approximations of non-linearities.
However, precision below 2 bits still causes severe accuracy loss.
Quantization-aware training (QAT) addresses this by simulating quantization with Straight-Through Estimators (STE), enabling models to adapt during training.  
LLM-QAT~\cite{liu2023llm}, BitDistiller~\cite{du2024bitdistiller}, BitNet~\cite{wang2023bitnetscaling1bittransformers}, and TernaryCLIP~\cite{zhang2025ternaryclip} achieve strong low-bit robustness but require full-model updates.

\paragraph{Ternary Neural Networks (TNNs).} 
Ternary quantization restricts weights to $\{-1,0,1\}$, offering large reductions in memory and compute by replacing multiplications with sign operations.  
Early work in CNNs, such as TWN~\cite{li2016ternary} and TTQ~\cite{zhu2016trained}, demonstrated that ternary weights with learned scaling factors are efficient and can retain accuracy, and later studies extended these ideas to transformers and LLMs~\cite{ma2024era}.  
Training ternary models is challenging due to the discrete nature of weights—updates must respect the quantization constraints of $\{-1,0,1\}$ rather than continuous values.  
Most methods rely on straight-through estimators and normalization or scaling strategies to stabilize optimization~\cite{deng2018gxnor}.  
Despite these advances, adapting ternary models to downstream tasks remains challenging. 
Approaches based on continuous, additive updates cause the merged weights to leave the ternary domain and miss out on the memory and compute benefits of ternary quantization. 
The need for ternary domain preservation motivates new designs that operate natively in the ternary space. Our adaptation applies low-rank updates to ternary base weights such that the result remains ternary.

\paragraph{PEFT Methods.}
Parameter-Efficient Fine-Tuning (PEFT) methods include adapters \cite{houlsby2019parameter,pfeiffer2021adapterfusion}, prefix/prompt tuning~\cite{li2021prefix,lester2021power}, and bias/activation-scaling~\cite{zaken2022bitfit,liu2022few}, which reduce trainable parameters but add extra modules or token overhead at inference. 
We focus on low-rank adaptation (LoRA)~\cite{hu2022lora}, which applies low-rank updates to weight matrices without altering model architecture or input tokens, making it naturally compatible with quantized models.

\paragraph{Low Rank Adaptation (LoRA).} LoRA~\cite{hu2022lora} decomposes weight updates into small low-rank matrices that are trained, while the base weights remain frozen. At inference, the low-rank matrices are merged with the base weights for adaptation without increasing inference cost. Most LoRA variants use additive updates, reducing trainable parameters while maintaining near full-model fine-tuning performance.  
Other additive approaches, like AdaLoRA~\cite{zhang2023adalora} and ReLoRA~\cite{lialin2023relora}, improve flexibility through adaptive budgets or periodic high-rank updates. Low-rank multiplicative adaptation (LoRMA)~\cite{bihany2025lorma} replaces additive updates with a more expressive framework based on multiplicative matrix transformations to rotate the weight space rather than shift it, and element-wise multiplicative updates, as in Fast LoRA (FLoRA)~\cite{wen2023batched}, compute input-specific modifications via Hadamard products.
An element-wise multiplicative operation between ternary base weights and a ternary adaptation matrix enables discrete ternary updates—flipping or zeroing weights—that preserve the ternary domain.

\paragraph{Low-rank Decomposition.} LoRA-based fine-tuning methods can be categorized into different types based on how they decompose the weight-update matrix. Simple low-rank decomposition of LoRA~\cite{hu2022lora} represents the update into two small matrices, reducing rank and parameters at the cost of expressivity. SVD-based decomposition dynamically allocates rank across layers by truncating or weighting singular values, as in AdaLoRA~\cite{zhang2023adalora}, SaLoRA~\cite{hu2023structure}, and IncreLoRA~\cite{zhang2023increlora}, improving flexibility while controlling parameter usage. Kronecker-product decomposition represents updates as block matrices via Kronecker products~\cite{graham2018kronecker}, preserving the effective rank of the original weight matrix with fewer parameters, as in LoKr~\cite{yeh2023navigating}, KAdaptation~\cite{he2023parameter} and KronA~\cite{edalati2022krona}. 
Similarly, KnGPT2~\cite{edalati2022kronecker} uses Kronecker decomposition to compress the original full-precision weights by representing them with smaller matrices.
Inspired by previous Kronecker-product decompositions, our ternary multiplicative adapter represents discrete ternary updates using two small ternary matrices, enabling high-rank update while preserving the ternary domain for flips and zeroing operations. 

\paragraph{PEFT with quantized backbone.} Recent adaptation methods extend low-bit fine-tuning to quantized large models.
QLoRA~\cite{dettmers2023qlora} combines LoRA with low-bit quantization by storing the base weights in a quantized format and dequantizing them to apply the full precision low-rank updates, which adds compute overhead but lowers memory usage.
QA-LoRA~\cite{xu2023qa} avoids dequantization for 2-bit quantized weights by merging the adaptation into per-group quantization parameters, but cannot enforce the merged weights to remain in the ternary domain.
QAT-based methods such as LoftQ~\cite{li2023loftq}, L4Q~\cite{jeon2024l4q}, and ApiQ~\cite{liao2024apiq} integrate quantization into training for improved accuracy. 
Yet all these approaches rely on continuous additive updates, which are incompatible with discrete ternary weights. Our method keeps the ternary domain even after the adaptation is fused with the base weights.

\section{Method}
\label{sec:method}

\begin{table*}[t]
    \centering
    \small
    \setlength{\tabcolsep}{1mm}
    \renewcommand{\arraystretch}{2}
    \caption{LoRA, QLoRA, QA-LoRA and Ours. Each column specifies the frozen base weights, the trainable low-rank adaptation, and the merged weights.
    LoRA keeps a full-precision base matrix $\mathbf{W}_{\text{0}}$ and adds a low-rank update $\mathbf{\Delta W} = \frac{\alpha}{r} \mathbf{B} \mathbf{A}$ in floating point.
    QLoRa stores a per-tensor quantized weight $\mathbf{W}_{\text{q}}$, dequantizes it, then adds the same low-rank update in full precision.
    QA-LoRA assumes group-wise affine quantization $\mathbf{W}_{\text{0,g}} = s_g (\mathbf{Q}_{\text{g}} - \mathbf{Z}_{\text{g}})$ and applies the low-rank update to the zero points $\mathbf{Z}_{\text{g}}$; using group size $1$ reduces to a per-tensor quantization.
    Our method starts from ternary base weights $\mathbf{W}_{\text{tern}} \in \{-1,0,1\}^{d_{\text{out}}  \times d_{\text{in}}}$, represents the adaptation as a Kronecker product, and merges it multiplicatively, so the final weights remain ternary.
    The first three methods rely on additive updates in a continuous space, whereas our method keeps the same weight space, such that the merged model stays ternary without needing any dequantization or requantization. }
    \resizebox{1\linewidth}{!}{%
    \begin{tabular}{l|l|l|l}
    \hline
    \textbf{Method} & \textbf{Base Weight (frozen)} & \textbf{Trainable Update} & \textbf{Merged Weight } \\
    \hline
    \textbf{LoRA}   & $\mathbf{W}_\text{0}$ (FP)    & $\mathbf{\Delta W} = \frac{\alpha}{r} \mathbf{B} \mathbf{A}$ & 
    $\begin{aligned} \mathbf{W_{\text{merged}}} &= \mathbf{W}_\text{0} + \frac{\alpha}{r} \mathbf{B} \mathbf{A}, \end{aligned}$ FP \\
    \hline
    \textbf{QLoRA} & Per-tensor quantized $\mathbf{W}_\text{q}$ & $\mathbf{\Delta W} = \frac{\alpha}{r} \mathbf{B} \mathbf{A}$ & 
    $\begin{aligned} \mathbf{W_{\text{merged}}} &= \text{Dequant}(\mathbf{W}_\text{q}) + \frac{\alpha}{r} \mathbf{B} \mathbf{A}, \end{aligned}$ FP \\
    \hline
    \textbf{QA-LoRA} & Group-wise quantized $\mathbf{W}_{\text{0,g}} = s_g (\mathbf{Q}_{\text{g}} - \mathbf{Z}_{\text{g}})$ &
    $\begin{aligned} &\text{Low-rank update to zero-point matrix:} \\ &\mathbf{\Delta Z_g} = \frac{\alpha}{r} \mathbf{B} \mathbf{A} \end{aligned}$ 
    & 
    $\begin{aligned} \mathbf{W_{\text{merged,g}}} &= s_g \big(\mathbf{Q}_{\text{g}} - (\mathbf{Z}_{\text{g}} + \mathbf{\Delta Z}_{\text{g}})\big) \\ &= \mathbf{W}_{\text{0,g}} - s_g \, \mathbf{\Delta Z}_{\text{g}}, \end{aligned}$ INT4 \\
    \hline
    \textbf{Ours} & Ternary $\mathbf{W}_{\text{tern}} \in \{-1,0,1\}^{d_{\text{out}}  \times d_{\text{in}}}$ & 
    $\mathbf{\Delta W_{\text{tern}}} = \mathbf{A} \otimes \mathbf{B}$ & 
    $\begin{aligned} \mathbf{W_{\text{merged}}} &= \mathbf{W}_{\text{tern}} \odot (\mathbf{A} \otimes \mathbf{B}), \end{aligned}$ Ternary \\
    \hline
    \end{tabular}%
    }
    \label{tab:technical-comparisons}
\end{table*}

Standard low-rank adaptation methods rely on additive updates, yet additive updates are incompatible with discrete ternary weights because adding real-valued updates drives the entries outside the ternary domain, see Table~\ref{tab:technical-comparisons}. We propose a ternary multiplicative adapter that i) models ternary updates multiplicatively so the adapted weights remain in \(\{-1,0,1\}\), and ii) decomposes updates as a Kronecker product of two \emph{ternary} factors for compactness and high expressivity. 

\subsection{Ternary multiplicative adaptation}
\label{sec:multiplicative}

To preserve the ternary constraint of using only $\{-1, 0, 1\}$ values throughout training and merging, we adopt an element-wise multiplicative formulation
that can flip, zero, or retain base weights. 

Formally, let $\mathbf{W}_{\text{tern}} \in \{-1, 0, 1\}^{d_{\text{out}} \times d_{\text{in}}}$ represent a pretrained ternary weight matrix of a linear layer, where $d_{\text{out}}$, $d_{\text{in}}$ are the output and input dimensions, respectively.  
We define the adapted ternary weight matrix $\mathbf{W}'_{\text{tern}}$ as

\begin{equation}
\label{eq:adapted}
    \mathbf{W}'_{\text{tern}} = \mathbf{W}_{\text{tern}} \odot \mathbf{\Delta}_{\text{tern}},
\end{equation}
where ternary element \(\mathbf{\Delta}_{ij}\) behaves as:
\begin{equation}
    \mathbf{\Delta}_{ij} =
    \begin{cases}
    +1,& \text{retain } \mathbf{W}_{ij},\\
    0,& \text{zero out } \mathbf{W}_{ij},\\
    -1,& \text{flip sign of } \mathbf{W}_{ij}.
    \end{cases}
\end{equation}

This multiplicative rule guarantees closure over $\{-1,0,1\}$, so $\mathbf{W}'_{\text{tern}}$ remains ternary and can be deployed without any floating-point operation, with the remark that, if $\mathbf{W}_{ij}=0$, then $\mathbf{W'}_{ij}=0$ for any $\mathbf{\Delta}_{ij}$, meaning that the multiplicative update cannot reactivate weights pruned to zero.
\subsection{Low-rank Kronecker decomposition}
\label{sec:kron}

Classical matrix decompositions—such as singular value decomposition (SVD), eigenvalue decomposition (EVD), QR factorization, CUR decomposition, and nonnegative matrix factorization (NMF)—cannot model discrete, sign-flipping ternary updates \(\mathbf{\Delta}_{\text{tern}}\) without breaking ternary domain constraints. Representing such discrete ternary updates in trainable low-rank matrices is difficult because the update space is both discrete and non-linear. 
Instead, we use a Kronecker factorization of \(\mathbf{\Delta}_{\text{tern}}\) into two smaller ternary matrices:

\begin{equation}  
    \mathbf{\Delta}_{\text{tern}} = \mathbf{A} \otimes \mathbf{B},
\end{equation}
where $\otimes$ denotes the Kronecker product between
\begin{equation}
    \mathbf{A} \in \{-1,0,1\}^{p\times q}, \qquad \mathbf{B} \in \{-1,0,1\}^{r\times s},
\end{equation}
which builds the matrix \(\mathbf{\Delta}_{\text{tern}}\) by replacing each entry of $\mathbf{A}$ with that entry multiplied by the entire matrix $\mathbf{B}$, see Figure~\ref{fig:ternary_adapter}.
The dimensions of $\mathbf{A}$ and $\mathbf{B}$ are defined by $p, q, r, s$, which are smaller than $d_{\text{out}}$, $d_{\text{in}}$ dimensions of the full weight matrix ${\bf W}_{\text{tern}}$, such that their pair-wise product
\begin{equation}
    p \cdot r = d_{\text{out}}, \qquad q \cdot s = d_{\text{in}}
    \label{eq:shape-match}
\end{equation}
matches the base weight shape \(({d_{\text{out}},d_{\text{in}}})\). By construction, the Kronecker product produces a ternary matrix \(\mathbf{\Delta}_{\text{tern}}\in\{-1,0,1\}^{{d_{\text{out}}  \times d_{\text{in}}}}\), so the element-wise product \(\mathbf{W}_{\text{tern}}\odot\mathbf{\Delta}_{\text{tern}}\) remains strictly ternary and can be merged directly without dequantization.

At the same time, the Kronecker product creates a high-rank update from two smaller matrices:
\begin{equation}
    \mathrm{rank}(\mathbf{A}\otimes \mathbf{B})=\mathrm{rank}(\mathbf{A}) \cdot \mathrm{rank}(\mathbf{B}) \le \min(p,q)\,\min(r,s) \le \min(d_{\text{out}}, d_{\text{in}}),
\end{equation}
while keeping the number of trainable parameters smaller 
than a full weight update.

\subsection{Training via real-valued proxies and STE}
Directly optimizing discrete \(\mathbf{A},\mathbf{B}\) is difficult; we therefore optimize real-valued proxies \(\bar{\mathbf{A}}\in\mathbb{R}^{p\times q},\bar{\mathbf{B}}\in\mathbb{R}^{r\times s}\) and obtain ternary factors by projection in the forward pass:
\begin{equation}
    \mathbf{A}=\operatorname{Tern}(\bar{\mathbf{A}}),\qquad \mathbf{B}=\operatorname{Tern}(\bar{\mathbf{B}}),
\end{equation}
where \(\operatorname{Tern}(\cdot)\) maps entries to \(\{-1,0,1\}\), with thresholds calculated from the absolute mean of each matrix.
Gradients are passed to the proxies using a straight-through estimator.

\paragraph{Inference.} 
After fine-tuning, we merge the base weights $\mathbf{W}_{\text{tern}}$ with the adaptation $\mathbf{A}\otimes \mathbf{B}$ by element-wise multiplication:
\begin{equation}
    \mathbf{W}_{\text{merged}} = \mathbf{W'}_{\text{tern}} =\mathbf{W}_{\text{tern}}\odot(\mathbf{A}\otimes \mathbf{B}).
\end{equation}
Since the final merged weights are ternary, \(\mathbf{W}_{\text{merged}}\in\{-1,0,1\}\), the adaptation weights $\bar{\mathbf{A}},\bar{\mathbf{B}}$ can be discarded, and the deployed layer is identical to the original ternary layer in terms of shape, precision and activations.

\subsection{Efficiency and expressivity}
\label{sec:analysis}

\paragraph{Trainable Parameters.} 
The Kronecker adaptation $\mathbf{\Delta}_{\text{tern}}=\mathbf{A}\otimes \mathbf{B}$ decomposes to shape $(p \cdot r)\times(q \cdot s)$. The number of trainable  parameters for a linear layer is
\begin{equation}
    n_{\text{params}} = p \cdot q + r \cdot s,
\end{equation}
which is typically much smaller than the  ${d_{\text{out}} \cdot d_{\text{in}}}$ parameters of a full update when $p, r, q ,s$ are chosen to be small factors of $d_{\text{out}}$ and $d_{\text{in}}$.

\noindent
\paragraph{Shape match and factor choice.} To match the shape constraints in Eq.~\eqref{eq:shape-match}, we choose $p, r$ such that $p \cdot r = d_{\text{out}}$ and $q, s$ such that $q \cdot s = d_{\text{in}}$. For square layers with $d_{\text{out}} = d_{\text{in}} = d$, the dimensions simplify to:
\begin{equation}
    p = r = q = s = \sqrt{d},
\end{equation}
which totals 
\begin{equation}
    n_{\text{params}} = 2 \cdot \sqrt{d} \cdot \sqrt{d} = 2d.
\end{equation}
In practice, transformer layer dimensions are large powers of $2$, so our factor choices can keep the Kronecker factors balanced and avoid bottlenecks from small factors.

\paragraph{Memory.} 
Table~\ref{tab:memory} reports per-layer fine-tuning memory. With FP32 real-valued proxies \(\bar{\mathbf{A}}\in\mathbb{R}^{p\times q},\bar{\mathbf{B}}\in\mathbb{R}^{r\times s}\),
each trainable parameter and its gradient occupy 4\,Bytes and Adam contributes 8\,Bytes per parameter. Consequently the adaptation for a square layer with ${d_{\text{out}}=d_{\text{in}}}=d$ uses $8d$\,Bytes (params), $8d$\,Bytes (grads), and $16d$\,Bytes (optimizer) per layer, whereas LoRA of rank $r$ uses $8dr$, $8dr$, and $16dr$\,Bytes, respectively. Matching our expressivity would require LoRA $r{=}d$ and therefore quadratic $O(d^{2})$ adaptation memory, while ours provides a similarly expressive update with only $O(d)$ memory.  

\paragraph{FLOPs.}
The forward pass for our method $\mathbf{W}_{\text{tern}}\odot(\mathbf{A}\otimes \mathbf{B})$ on a square layer with ${d_{\text{out}}=d_{\text{in}}}=d$ costs $2d^2$ operations, which matches the 
full weights update cost, while LoRA’s forward pass amounts to more, $d^2(2r+1)$. However, after training, at inference time, both methods incur zero extra FLOPs: LoRA parameters are merged via $\mathbf{W} + \mathbf{BA}$, and ours via $\mathbf{W}_{\text{tern}}\odot(\mathbf{A}\otimes \mathbf{B})$, both producing a single $\mathbf{W}_{\text{merged}}$ reused across all tokens. The inference-time adapter overhead is therefore zero. Our method has the important advantage of ternary domain preservation after merging. 

\begin{table}[t]
\setlength{\tabcolsep}{0.2mm}
\centering
\caption{Per-layer training memory, Bytes (B). Adapters are FP32: 4\,B/param, 4\,B/grad, 8\,B/Adam. Quantized versions are recomputed during each forward pass, gradients are computed using a straight-through estimator. Activations are FP16 at 2\,B/element. For a $\mathbf{W}\!\in\!\mathbb{R}^{d\times d}$; \textit{General}: $\bar{\mathbf{A}}\!\in\!\mathbb{R}^{p\times q}$, $\bar{\mathbf{B}}\!\in\!\mathbb{R}^{r\times s}$ s.t. $p\!\cdot\!r\!=\!q\!\cdot\!s\!=\!d$. \textit{Square}: $p\!=\!q\!=\!r\!=\!s\!=\!\sqrt{d}$. Our adaptation scales as $O(d)$ memory in the square case, versus $O(d^2)$ for full fine-tuning and $O(dr)$ for LoRA; matching our expressivity would require LoRA rank $r\!=\!d$.}
\label{tab:memory}
\small
\begin{tabular*}{\columnwidth}{@{\extracolsep{\fill}}lcccc}
\toprule
\textbf{Method} & \textbf{Train.\ params.} & \textbf{Grads.} & \textbf{Opt.} & \textbf{Acts.} \\
\midrule
Full-FT (FP32) & $4d^{2}$ & $4d^{2}$ & $8d^{2}$ & $2\,bsd$ \\
LoRA (rank $r$) & $8dr$ & $8dr$ & $16dr$ & $2\,bsd{+}2\,bsr$ \\
Ours (general) & $4(pq{+}rs)$ & $4(pq{+}rs)$ & $8(pq{+}rs)$ & $2\,bsd$ \\
Ours (square) & $8d$ & $8d$ & $16d$ & $2\,bsd$ \\
\bottomrule
\end{tabular*}
\end{table}
\section{Experiments}
\label{sec:experiments}

We empirically evaluate our ternary multiplicative adaptation on (i) two ternarized LLMs, (ii) three pre-trained ternary LLMs, and (iii) a ternary ViT.

\subsection{Experimental Setup}

\paragraph{Models and baselines.}
We consider three evaluation tracks.

\textit{(i) Ternary PTQ LLM backbones.} We use Llama-3.2-1B~\cite{meta_llama_3_2_1b_2024} and Llama-3.2-3B~\cite{meta_llama_3_2_3b_2024} as base models. The weights are first quantized to ternary using SpinQuant~\cite{liu2024spinquant}, which creates the ternary backbones in which we insert our adaptation. The base ternary weights are stored in UINT8 packed format to reduce memory usage, with corresponding scales stored in BFloat16 precision. The baselines are: full precision (FP, 16-bit), 2-bit Round-To-Nearest (RTN), 2-bit GPTQ~\cite{frantar2022gptq}, 2-bit SpinQuant~\cite{liu2024spinquant}, and ternary SpinQuant~\cite{liu2024spinquant}.
We additionally compare against QLoRA~\cite{dettmers2023qlora}, as a baseline for existing PEFT applied on ternary backbones.

\textit{(ii) Pre-trained ternary backbones.} We fine-tune the following pre-trained, ternary models: Falcon-E-1B-Instruct~\cite{tiionebitllms}, Falcon-E-3B-Instruct~\cite{tiionebitllms}, and BitNet b1.58 2B4T~\cite{ma2025bitnet}. We compare our fine-tuned adaptations against the corresponding non-fine-tuned version.

\textit{(iii) Ternary ViT backbone.} We use our method to fine-tune a ternary ViT-B/16 (vision encoder from TernaryCLIP~\cite{zhang2025ternaryclip}) and compare against full ternary fine-tuning and QLoRA baselines.

\paragraph{Datasets and evaluation metrics.}
For the PTQ LLM backbones we fine-tune on Alpaca dataset~\cite{taori2023stanford} and evaluate on the following task set: ARC-Challenge (ARC-c) and ARC-Easy (ARC-e)~\cite{clark2018think}, BoolQ~\cite{clark2019boolq}, CommonsenseQA (ComQA)~\cite{talmor2018commonsenseqa}, HellaSwag~\cite{zellers2019hellaswag}, MMLU~\cite{hendrycks2020measuring}, OpenBookQA (OBQA)~\cite{mihaylov2018can}, PiQA~\cite{bisk2020piqa}, Winogrande (Wino.)~\cite{sakaguchi2021winogrande}, and WikiText-2~\cite{merity2016pointer} token-level perplexity (PPL). All evaluations are zero-shot and use the default \texttt{lm-evaluation-harness} (v0.4.9.1) per-task configurations.
For pre-trained ternary backbones we fine-tune on GSM8K~\cite{cobbe2021training} and report exact-match (EM) accuracy on the test split, using each model's chat template, system instruction ("You are a helpful assistant"), and 4, 5-fewshots, where the fewshot examples are treated as a multi-turn conversation in \texttt{lm-evaluation-harness} (v0.4.9.1).
For the ViT, we fine-tune on ImageNet-100~\cite{ambityga_imagenet100_kaggle} and report Top-1 classification accuracy.

\paragraph{Adapter configuration and training details.} 
We apply our ternary multiplicative adaptation to all self-attention and feed-forward blocks in the Transformer, excluding the task-specific output head (the language model head for LLMs and the classification head for ViTs).
For a weight matrix $\mathbf{W}_{\text{tern}} \in \mathbb{R}^{m \times n}$, we choose factor dimensions $(p,q)$ and $(r,s)$ such that $m = p \cdot r$ and $n = q \cdot s$. We select $p,q,r,s$ to be as balanced as possible under the integer divisibility constraints. For square layers we use $p = q = r = s = \sqrt{d}$. Table~\ref{tab:factorization} lists the factorization for all Llama-3.2-1B layer shapes, yielding $0.06\%$ trainable parameters. SpinQuant~\cite{liu2024spinquant} quantization uses per-channel absmax symmetric quantization on 800 WikiText-2 samples. Activations remain in the original precision, 16-bit for Llama and ViT, and 8-bit for BitNet and Falcon. We optimize the real-valued proxies of the ternary factors with a straight-through estimator as described in \S\ref{sec:multiplicative}.
We fine-tune the PTQ backbones for one epoch using AdamW with learning rate $1.5 \times 10^{-3}$, a linear decay schedule, warmup ratio $0.03$, on-device batch size $16$, on a single NVIDIA A40 GPU (48GB). Learning rate is $1.0 \times 10^{-4}$ for the pre-trained ternary backbones. The supplementary material reports more details about the training dynamics of our method.

\paragraph{Initialization.}
\label{par:init}
We initialize real-valued proxies $\bar{\mathbf{A}},\bar{\mathbf{B}}$ so that, at the start of fine-tuning, the adapted ternary weight matrix $\mathbf{W}'_{\text{tern}}$ from Eq.~\eqref{eq:adapted} used in the forward pass is identical to the pre-trained $\mathbf{W}_{\text{tern}}$. We use three strategies:
(i) \emph{All-ones}: set every entry of $\bar{\mathbf{A}}$ and $\bar{\mathbf{B}}$ to $+1$, yielding a multiplicative identity update;
(ii) \emph{Balanced}: fill $\bar{\mathbf{A}}$ and $\bar{\mathbf{B}}$ with an equal number of $+1$ and $-1$ values, and compensate the base weights via $\mathbf{W}_{\text{tern}} \leftarrow \bigl( \operatorname{sign}\bar{\mathbf{A}} \otimes \operatorname{sign}\bar{\mathbf{B}} \bigr) \odot \mathbf{W}_{\text{tern}}$;
(iii) \emph{Normalized}: sample each entry in $\bar{\mathbf{A}}$ and $\bar{\mathbf{B}}$ from $u \sim \mathcal{U}(0.6, 1.4)$ with random signs, then normalize to have mean absolute value $1$, and apply the same sign compensation as in (ii).
We use \textit{Balanced} for the PTQ LLM and ViT backbones and \textit{Normalized} for the GSM8K experiments. \textit{All-ones} is used only in the analysis of weight transformations. The supplementary material reports the corresponding ablation for the three strategies.

\begin{table*}[ht]
\centering
\caption{Main results on Llama 3.2-1B and Llama 3.2-3B. Our 1.58-bit method is compared against full-precision (FP) and several post-training quantization (PTQ) baselines. All evaluations are zero-shot. Best results among 1.58-bit methods are in \textbf{bold}. Averages exclude WikiText-2 PPL ($\downarrow$). Our ternary multiplicative adaptation consistently recovers much of the accuracy lost to ternarization and often matches or surpasses 2-bit baselines, while roughly halving the PPL of the ternary SpinQuant backbones.}
\label{tab:main_results}
\resizebox{\textwidth}{!}{
\begin{tabular}{llcccccccccccc}
\toprule
\toprule
\textbf{Model} & \textbf{Method} & \textbf{Precision} & \textbf{ARC-c} & \textbf{ARC-e} &
\textbf{BoolQ} & \textbf{ComQA} & \textbf{HellaSwag} & \textbf{MMLU} & \textbf{OBQA} &
\textbf{PiQA} & \textbf{Wino.} & \textbf{Avg.} & \textbf{WikiText-2 PPL $\downarrow$} \\
\midrule
\multirow{7}{*}{\textbf{Llama 3.2-1B}}
  & Full Precision & 16-bit & 31.4 & 65.2 & 63.6 & 47.0 & 47.7 & 36.7 & 26.4 & 74.6 & 59.8 & 50.2 & 9.7 \\
\cmidrule(lr){2-14}
  & \multicolumn{13}{l}{\textit{PTQ Baselines}} \\
  & RTN       & 2-bit   & 22.8 & 23.8 & 62.1 & 19.2 & 25.5 & 22.9 & 17.2 & 53.1 & 48.9 & 32.8 & 1.5e6 \\
  & GPTQ      & 2-bit   & 20.2 & 31.6 & 43.1 & 19.3 & 26.3 & 24.1 & 13.4 & 55.2 & 50.0 & 31.4 & 1.7e2 \\
  & SpinQuant & 2-bit   & 18.3 & 37.6 & 62.0 & 19.5 & 28.7 & 22.9 & 13.4 & 57.4 & 53.2 & 34.7 & 43.3 \\
  & SpinQuant & 1.58-bit& 20.3 & 29.7 & \textbf{61.3} & \textbf{20.1} & 27.0 & 22.9 & 13.8 & 54.5 & 49.2 & 33.2 & 86.5 \\
\cmidrule(lr){2-14}
  & \textbf{Ours (Adapter)} & 1.58-bit & 19.5 & \textbf{39.7} & 61.2 & 19.9 &
    \textbf{29.4} & \textbf{23.1} & 13.8 & \textbf{58.2} & \textbf{53.2} & \textbf{35.3} & \textbf{44.6} \\
\midrule
\multirow{7}{*}{\textbf{Llama 3.2-3B}}
  & Full Precision & 16-bit & 42.4 & 74.6 & 72.9 & 64.0 & 55.2 & 54.0 & 31.0 & 76.6 & 69.5 & 60.0 & 7.8 \\
\cmidrule(lr){2-14}
  & \multicolumn{13}{l}{\textit{PTQ Baselines}} \\
  & RTN       & 2-bit   & 22.5 & 24.4 & 37.8 & 19.2 & 25.3 & 26.8 & 15.2 & 52.6 & 50.2 & 30.4 & 7.6e5 \\
  & GPTQ      & 2-bit   & 21.5 & 25.2 & 40.8 & 19.2 & 26.0 & 22.9 & 13.2 & 53.1 & 48.4 & 30.0 & 1.9e2 \\
  & SpinQuant & 2-bit   & 19.3 & 29.3 & 53.6 & 19.7 & 28.2 & 22.9 & 15.4 & 57.0 & 48.7 & 32.6 & 41.7 \\
  & SpinQuant & 1.58-bit& 18.4 & 33.3 & 45.6 & 19.5 & 27.9 & 22.9 & 11.6 & 57.2 & 51.2 & 31.9 & 45.6 \\
\cmidrule(lr){2-14}
  & \textbf{Ours (Adapter)} & 1.58-bit & \textbf{23.8} & \textbf{46.0} & \textbf{62.2} & \textbf{20.3} &
    \textbf{34.0} & \textbf{24.8} & \textbf{16.4} & \textbf{64.6} & \textbf{52.7} & \textbf{38.3} & \textbf{22.3} \\
\bottomrule
\bottomrule
\end{tabular}}
\end{table*}

\begin{table}[ht]
\caption{Balanced factorization for self-attention and feed-forward Neural Network weights shapes (excluding the $lm\_head$) for Llama-3.2-1B layers. The factors $(p,q)$ and $(r,s)$ satisfy $m{=}p\!\cdot\!r$ and $n{=}q\!\cdot\!s$, yielding balanced submatrices for all layer shapes. This factorization keeps the number of trainable adapter parameters tiny (about $0.06\%$ of the model) while still allowing the multiplicative update to cover every weight in each layer, enabling expressive adaptations at small parameter cost.}
\label{tab:factorization}
\setlength{\tabcolsep}{4pt}
\centering
\footnotesize
\begin{tabular}{@{}l @{\;\,$\to$\;\;} l@{}}
\toprule
\textbf{Shape $\mathbf{W}_{\text{tern}}$ ($m\times n$)} & \textbf{Adaptation dims $(p\!\times\!q,\ r\!\times\!s)$} \\
\midrule
$(2048\times 2048)$ & $(32\!\times\!32,\ 64\!\times\!64)$ \\
$(2048\times 512)$  & $(32\!\times\!16,\ 64\!\times\!32)$ \\
$(2048\times 8192)$ & $(32\!\times\!64,\ 64\!\times\!128)$ \\
$(8192\times 2048)$ & $(64\!\times\!32,\ 128\!\times\!64)$ \\
\bottomrule
\end{tabular}

\end{table}

\subsection{Evaluations on LLM PTQ Backbones} 
Table \ref{tab:main_results} reports results for Llama-3.2-1B. Relative to the direct ternary SpinQuant baseline, our adaptation improves performance on six out of nine benchmarks, while being equal or at most 0.8 percentage points worse on the other three, raising the average score from $33.2$ to $35.3$. The PPL is improved from $86.5$ to $44.6$. We surpass the 2-bit SpinQuant baseline on seven out of nine benchmarks and almost match its $43.3$ PPL, while retaining the ternary domain. 

Table \ref{tab:main_results} shows a stronger trend for Llama-3.2-3B: our method improves upon the ternary SpinQuant baseline on all nine benchmarks, raising the average accuracy from $31.9$ to $38.3$ and reducing PPL from $45.6$ to $22.3$. It also outperforms the 2-bit SpinQuant baseline on every benchmark while lowering PPL from $41.7$ to $22.3$. 

Because our adaptation is merged into the base ternary weights and the result is a still ternary model, these accuracy gains are obtained without additional parameters or latency at inference time, unlike having a separate full-precision adaptation (e.g., in QLoRA), which adds runtime overhead.

\subsection{Comparison to QLoRA baseline (LLM)}

We fine-tune QLoRA on the same Llama-3.2-3B ternary backbone and Alpaca dataset. Table~\ref{tab:qlora_finetunes} compares our method to QLoRA merged and requantized to 1.58-bit, showing that requantization degrades accuracy (38.3 vs. 37.5) and PPL (22.3 vs 22.9), whereas our method works natively and achieves better performance. 

\begin{table}[t]
    \centering
    \small
    \caption{Llama-3.2-3B fine-tuning results on Alpaca dataset, compared to QLoRA requantization, showing average accuracy over 9 tasks and WikiText-2 perplexity. \textit{QLoRA (requantized)} fine-tunes with full-precision adapters but then merges and requantizes the resulting weights back to ternary, incurring a drop in accuracy and PPL, whereas our method works natively in the ternary domain and achieves better performance. }
    \label{tab:qlora_finetunes}
    \setlength{\tabcolsep}{5pt}
    \begin{tabular}{lccc}
    \toprule
    \textbf{Method} & \textbf{Bits (Base/Adapter)} & \textbf{Avg.\ $\uparrow$} & \textbf{PPL $\downarrow$} \\
    \midrule
    QLoRA (requantized)  & 1.58 / 1.58 & 37.5 & 22.9 \\
    Ours (merged ternary)   & 1.58 / 1.58 & 38.3 & 22.3 \\
    \bottomrule
    \end{tabular}
\end{table}

\subsection{Downstream Task Adaptation (LLM)} 
\label{sec:downstream}
We evaluate whether our method can adapt pre-trained ternary models to a downstream task. We use Falcon-E-1B-Instruct, Falcon-E-3B-Instruct~\cite{tiionebitllms}, and BitNet b1.58 2B4T~\cite{ma2025bitnet} as pre-trained ternary backbones, and fine-tune only our adaptation's parameters on GSM8K~\cite{cobbe2021training}. Table~\ref{tab:gsm8k} summarizes the results. Our adaptation consistently improves over the corresponding ternary baselines, with gains of $+2.9$, $+3.1$, and $+1.0$ EM points, respectively. These improvements indicate that strongly quantized backbones can be specialized with a tiny number of trainable parameters while keeping the merged model strictly ternary and incurring no inference overhead.

\begin{table}[t]
\centering
\small
\caption{GSM8K exact-match accuracy (EM, \%) for pre-trained ternary models fine-tuned with our adaptation. 
All models use 1.58-bit weights and are fine-tuned on GSM8K with only the adaptation parameters updated. Overall, our method improves $+1$ to $+3$ EM points over the frozen ternary backbones, demonstrating that strongly quantized models can be adapted to downstream math reasoning tasks without increasing inference-time cost.}
\label{tab:gsm8k}
\begin{tabular}{lcccc}
\toprule
Model & \#fewshot & Baseline & Ours & Improvement \\
\midrule
BitNet 2B 4T      & 4 & 60.1 & \textbf{63.0} & +2.9 \\
Falcon Edge 1B    & 5 & 52.0 & \textbf{55.1} & +3.1 \\
Falcon Edge 3B    & 5 & 65.4 & \textbf{66.4} & +1.0 \\
\bottomrule
\end{tabular}
\end{table}

\subsection{Evaluations on ViT}
Table~\ref{tab:vit_finetunes} reports ImageNet-100 fine-tuning results for a ternary ViT-B/16 (from TernaryCLIP~\cite{zhang2025ternaryclip}). We compare our method against full ternary fine-tuning, QLoRA with 16-bit adapters, and QLoRA requantized to 1.58-bit. Our method improves Top-1 accuracy over requantized QLoRA, while yielding a single merged 1.58-bit ViT model.

\begin{table}[t]
    \centering
    \small
    \caption{Ternary ViT-B/16 (encoder from TernaryCLIP~\cite{zhang2025ternaryclip}) fine-tuned on ImageNet-100. \textit{Full-FT} updates all ternary base weights with STE. \textit{QLoRA} trains full-precision adapters on the ternary backbone, and \textit{QLoRA (requantized)} additionally merges and requantizes to a single ternary ViT. While \textit{QLoRA} can match full fine-tuning, merging and requantizing causes a large accuracy drop. Our method avoids this requantization performance drop by fine-tuning and merging natively in the ternary domain.}
    \label{tab:vit_finetunes}
    \setlength{\tabcolsep}{5pt}
    \begin{tabular}{lcc}
    \toprule
    \textbf{Method} & \textbf{Bits (Base/Adapter)} & \textbf{Top-1 acc.\ $\uparrow$} \\
    \midrule
    Full-FT               & 1.58 & 85.7 \\
    QLoRA                 & 1.58 / 16 & 85.3 \\
    QLoRA (requantized)   & 1.58 / 1.58 & 78.9 \\
    Ours (merged ternary) & 1.58 / 1.58 & 83.0 \\
    \bottomrule
    \end{tabular}
\end{table}

\begin{table}[ht]
\centering
\caption{Element-wise ternary weight transitions from the SpinQuant Llama-3.2-1B backbone ($\mathbf{W}_\text{tern}$) to the merged model ($\mathbf{W}_\text{merged}$) after training, for different initialization schemes. Each row shows, in millions of weights, how many parameters with a given initial value (\emph{from}) end up at each final ternary state. The $0$ row is identical across all initializations because our multiplicative update cannot reactivate pruned ($0$) weights, so all accuracy gains come from redistributing signs among non-zero weights.}
\label{tab:ternary_transitions_counts}
\setlength{\tabcolsep}{5pt}
\small
\begin{tabular}{llccc}
\toprule
& & \multicolumn{3}{c}{Final value (millions of weights)} \\
\cmidrule(lr){3-5}
Init. & From & $-1$ & $0$ & $+1$ \\
\midrule
\multirow{3}{*}{All-ones}
& $-1$ & 255.1 & 4.1 & 0.0 \\
& $0$  & 0.0   & 454.8 & 0.0 \\
& $+1$ & 0.0   & 4.1 & 255.0 \\
\midrule
\multirow{3}{*}{Balanced}
& $-1$ & 127.5 & 4.1 & 127.5 \\
& $0$  & 0.0   & 454.8 & 0.0 \\
& $+1$ & 127.5 & 4.1 & 127.5 \\
\midrule
\multirow{3}{*}{Normalized}
& $-1$ & 126.6 & 6.0 & 126.6 \\
& $0$  & 0.0   & 454.8 & 0.0 \\
& $+1$ & 126.6 & 6.0 & 126.6 \\
\bottomrule
\end{tabular}
\end{table}

\subsection{Weight Transformations (LLM)} 
\label{sec:weight-transforms}
To understand how our adaptation recovers accuracy for the PTQ LLM backbones, we compare the base ternary weights $\mathbf{W}_\text{tern}$ produced by SpinQuant with the merged weights $\mathbf{W}_{\text{merged}}$ after fine-tuning and merging our adaptation. For Llama-3.2-1B, we compute element-wise transitions between ternary states $\{-1, 0, 1\}$ and report how many weights with a given initial value end up in each final state. We ablate the three initialization schemes from \S\ref{par:init}: \textit{All-ones}, \textit{Balanced}, and \textit{Normalized}, while keeping the training setup fixed.

Table~\ref{tab:ternary_transitions_counts} shows the raw transition counts for all $N = 973{,}078{,}528$ weights and expressed in millions of weights for readability. Here $N$ corresponds to all ternary weights in the self-attention and feed-forward projection matrices across the 16 Transformer layers of the model; the token embeddings, language model head, and normalization parameters remain in higher precision and are excluded from this analysis. In all cases, weights that are quantized to $0$ remain exactly zero after adaptation: no element transitions out of $0$. This reflects a structural property of our multiplicative update, which can only modulate the sign of non-zero ternary weights but cannot reactivate pruned connections. Consequently, any accuracy gains come from reassigning the signs of existing non-zero weights.
The supplementary material provides additional analysis of these weight transformations.

\section{Concluding Remarks}
\label{sec:conclusion}

We addressed parameter-efficient fine-tuning of ternary transformer backbones under a constraint that is difficult to satisfy with existing LoRA-style methods: the merged model should remain strictly in $\{-1,0,1\}$ without any dequantization or post-hoc requantization. To this end, we proposed a ternary multiplicative adapter that applies discrete keep/zero/flip updates directly to ternary weights, with the update parameterized by a low-rank Kronecker product of two small ternary factors. Across six models spanning language and vision, this design recovers a substantial portion of the performance lost to ternarization: it improves accuracy and perplexity over ternary PTQ baselines on ternarized LLaMA-3.2 1B/3B, outperforms requantized QLoRA under the same merged 1.58-bit constraint, and yields a single merged 1.58-bit ViT-B/16 that improves over requantized QLoRA on ImageNet-100 while narrowing the gap to full ternary fine-tuning. Finally, our weight-transition analysis shows that gains arise primarily from redistributing signs among non-zero weights. A limitation is that the multiplicative update cannot reactivate zero weights, since ternary weights have no zero-point to absorb the additive corrections. This zero-locking is the trade-off for strict ternary closure, and the Kronecker structure may constrain update patterns for poorly factorable layer shapes.

\section*{Acknowledgements}
This work is supported by the Dutch Research Council (NWO) through the Vici ENW project dAIta: Data Efficient AI Foundation Models, file number VI.C.242.088,
grant DOI: \url{https://doi.org/10.61686/KNPTR13127}. Experimental research reported in this work was facilitated by computational resources of the Delft AI Cluster (DAIC)~\cite{DAIC} at Delft University of Technology, The Netherlands.

\bibliographystyle{splncs04}
\bibliography{main}

\setcounter{section}{0}
\renewcommand{\thesection}{S\arabic{section}}
\renewcommand{\thesubsection}{\thesection.\arabic{subsection}}

\newpage
\section*{Supplementary Material}

\section{Additional analysis on weight transitions}
\label{sec:additional-weight-transitions}

\newcolumntype{C}[1]{>{\centering\arraybackslash}m{#1}}

\begin{table}[ht]
\centering
\caption{Summary of weight transformations for Llama-3.2-1B. Percentages are computed over all $N{=}973{,}078{,}528$ ternary weights. ``Unchanged'' counts entries whose initial and final values are identical. ``$-1\leftrightarrow+1$'' counts sign changes between the two non-zero states. ``NZ$\rightarrow 0$'' counts pruned non-zero weights. ``Flip in NZ'' reports the fraction of originally non-zero weights that flip sign after adaptation.}
\label{tab:ternary_transitions_summary}
\setlength{\tabcolsep}{4pt}
\small
\begin{tabular}{@{}l C{1.2cm} C{1.6cm} C{1.35cm} C{1.65cm}@{}}
\toprule
{\scriptsize Init.} &
{\scriptsize Unchanged\ (\%)} &
{\scriptsize $-1\leftrightarrow+1$\ (\%)} &
{\scriptsize NZ$\rightarrow 0$\ (\%)} &
{\scriptsize Flip in NZ (\%)} \\
\midrule
All-ones   & 99.16 & 0.00  & 0.84 & 0.00 \\
Balanced   & 72.95 & 26.21 & 0.85 & 49.21 \\
Normalized & 72.76 & 26.02 & 1.22 & 48.85 \\
\bottomrule
\end{tabular}
\end{table}

The three initialization strategies induce different patterns on the non-zero weights. With the \textit{all-ones} initialization, the merged model is very close to the original backbone: $99.16\%$ of all weights remain unchanged, see Table~\ref{tab:ternary_transitions_summary}. The only modifications are a small fraction $0.84\%$ of non-zero weights that are pruned to $0$. In particular, there are no direct sign flips between $-1$ and $+1$.

In contrast, both \textit{Balanced} and \textit{Normalized} initializations lead to significant sign flipping. Approximately $26\%$ of all weights flip between $-1$ and $1$, while less than $1\%$ of weights are pruned to zero. When restricted to non-zero weights, this corresponds to flipping about half of the active connections: $49.21\%$ for \textit{Balanced}, $48.85\%$ for \textit{Normalized}; see Table~\ref{tab:ternary_transitions_summary}. The two schemes yield very similar global statistics, suggesting that they primarily differ in optimization dynamics rather than in the final types of transformations they enable.

Overall, these results indicate that our ternary multiplicative adaptations act mainly by reorganizing the sign pattern of non-zero ternary weights, not by changing sparsity. The large number of sign flips under \textit{Balanced} and \textit{Normalized} initialization explains why these configurations are able to recover most of the accuracy lost to aggressive ternarization while keeping the final model strictly ternary.

\section{Initialization Ablation}
\label{sec:init-ablation}

We treat the initialization of the adaptation as a hyperparameter and evaluate the three initialization strategies described in the main paper on Llama-3.2-1B. Table~\ref{tab:init_ablation} shows that \textit{Balanced} gives the best average accuracy and lowest perplexity.

\begin{table}[h]
\centering
\small
\setlength{\tabcolsep}{4pt}
\caption{Initialization ablation on Llama-3.2-1B.}
\label{tab:init_ablation}
\begin{tabular}{lcc}
\toprule
Initialization & Avg. $\uparrow$ & PPL $\downarrow$ \\
\midrule
Balanced & 35.3 & 44.6 \\
All-ones & 35.1 & 45.7 \\
Normalized & 34.9 & 52.0 \\
\bottomrule
\end{tabular}
\end{table}

\section{Training Dynamics}
\label{sec:training-dynamics}

\begin{figure}[h]
    \centering
    \includegraphics[width=0.65\textwidth]{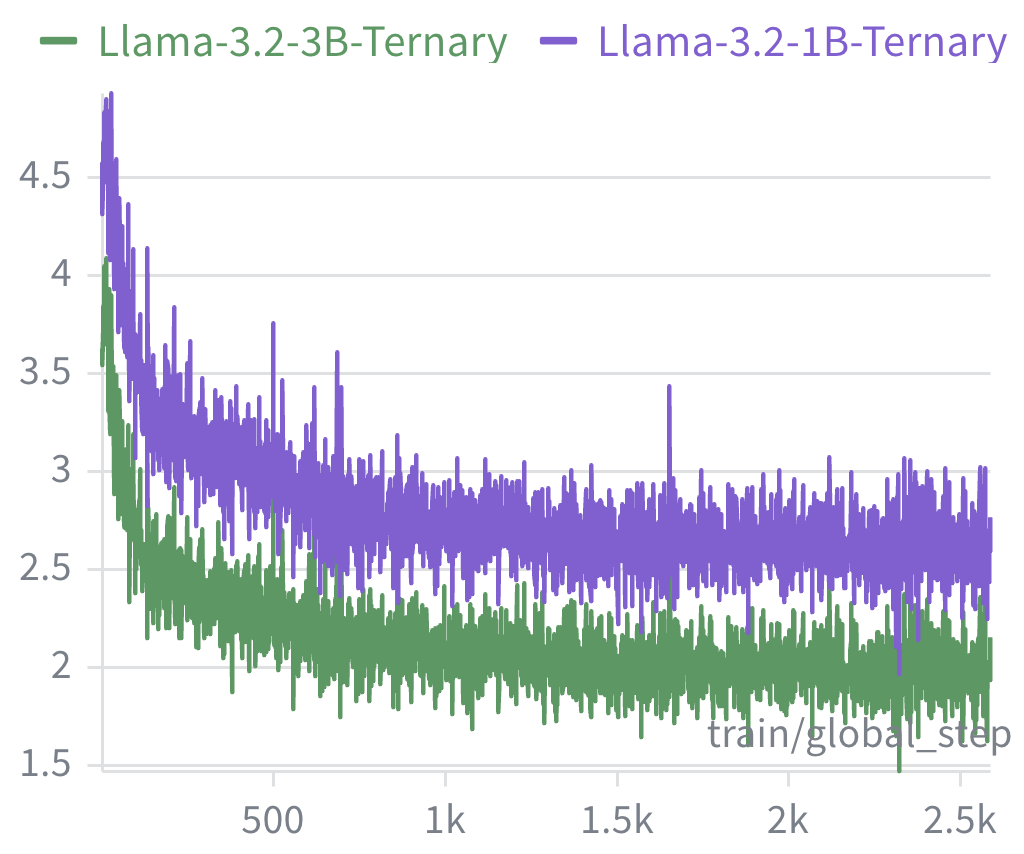}
    \caption{Training loss on Llama-3.2-1B.}
    \label{fig:loss}
\end{figure}

\begin{figure}[t]
    \centering
    \includegraphics[width=0.65\textwidth]{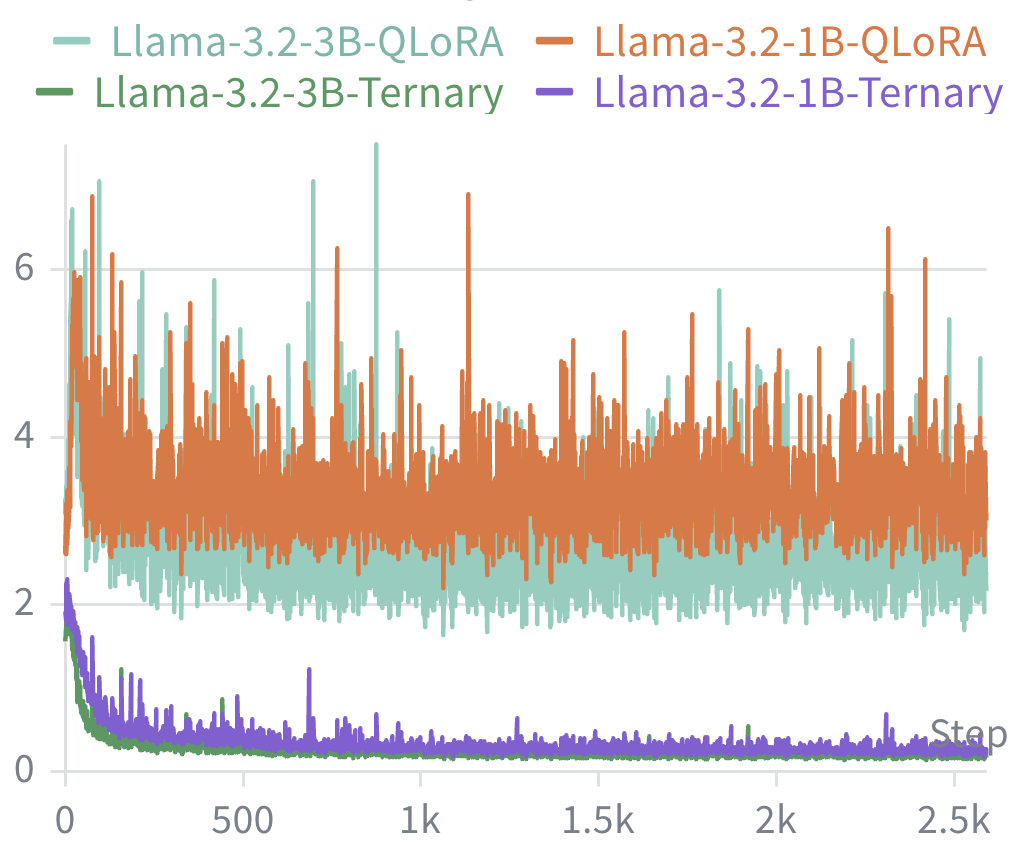}
    \caption{Gradient norm on Llama-3.2-1B.}
    \label{fig:grad_norm}
\end{figure}

On Llama-3.2-1B, our method reaches 31.0GB peak VRAM and completes one epoch in 1.43h, compared with 31.2GB and 1.83h for QLoRA~\cite{dettmers2023qlora}. Figure~\ref{fig:loss} and Figure~\ref{fig:grad_norm} show smooth loss convergence and stable gradient norms during training.

\end{document}